# Partial Labeled Gastric Tumor Segmentation via patch-based Reiterative Learning

**Yang Nan [a, b, c], Gianmarc Coppola [d], Qiaokang Liang [a, b, c, *], Kunglin Zou [a, b, c], Wei Sun [a, b, c], Dan Zhang [e], Senior Member, IEEE, Yaonan Wang [a, b], Guanzhen Yu [f, *]**

[a] College of Electrical and Information Engineering, Hunan University, Changsha 410082, China (e-mail: qiaokang@hnu.edu.cn)

[b] National Engineering Laboratory for Robot Vision Perception and Control, Hunan University, Changsha, Hunan 410082, China;

[c] Hunan Key Laboratory of Intelligent Robot Technology in Electronic Manufacturing, Hunan 418000, China;

[d] Faculty of Engineering and Applied Science, University of Ontario Institute of Technology, Oshawa, Ontario, L1H 7K4, Canada;

[e] Department of Mechanical Engineering, York University, Toronto, ON M3J 1P3, Canada;

[f] Department of oncology, Longhua Hospital affiliated to Shanghai University of Traditional Chinese Medicine, Shanghai 201203, China (e-mail: qiaoshanqian@aliyun.com)

## Abstract

Gastric cancer is the second leading cause of cancer-related deaths worldwide, and the major hurdle in biomedical image analysis is the determination of the cancer extent. This assignment has high clinical relevance and would generally require vast microscopic assessment by pathologists. Recent advances in deep learning have produced inspiring results on biomedical image segmentation, while its outcome is reliant on comprehensive annotation. This requires plenty of labor costs, for the ground truth must be annotated meticulously by pathologists. In this paper, a reiterative learning framework was presented to train our network on partial annotated biomedical images, and superior performance was achieved without any pre-trained or further manual annotation. We eliminate the boundary error of patch-based model through our overlapped region forecast algorithm. Through these advisable methods, a mean intersection over union coefficient (IOU) of 0.883 and mean accuracy of 91.09% on the partial labeled dataset was achieved, which made us win the 2017 China Big Data & Artificial Intelligence Innovation and Entrepreneurship Competitions[1].

## 1 Introduction

Treatment of cancer is an insurmountable gap for human beings. Data shows that the incidence of gastric cancer in the world are mainly in China, Japan and South Korea due to their unhealthy eating habits. Researchers have been experimenting with digital histopathology in recent years to assist pathologists in their diagnosis. With the improvement of computing capacity, deep learning has produced inspiring results on biomedical image segmentation and a series of supervised learning frameworks have been proposed[1][2][3][4]. However, methods with deep learning for biomedical segmentation still face a severe situation, i.e. the insufficiency data for training networks due to the labor costs. Compared with the annotation of the natural image segmentation problem, biomedical image segmentation data requires professional labeling and a great deal of patience. To overcome the absence of training data, several researchers have come up with semi-supervised learning to reduce the labor costs[5][6]. Nevertheless, most methods e.g. Ref. [2] are mainly aimed at natural images segmentation (such as "car", "person", "dog" or "horse") and are not suitable for solving biomedical image segmentation problems due to the coarse results. The appearance of active learning reduces the cost of sample annotating to a certain extent, but it requires a subset of precisely labeled datasets and further manual annotations[7][8].

[1] The challenge is held by Shanghai Big Data Alliance and Center for Applied Information Communication Technology (CAICT), the home page is http://www.datadreams.org/race-race-3.html

Recent advances in image segmentation have been devoted to promote architectural designs through the following approaches:

(1) Adding short skip connections[9][10] or altering base network[11][12][13] so as to promote the generalization ability of the network

(2) Expanding from 2D convolution to 3D convolution structures[14][15][16] in order to work out detection problems in CT images

(3) Adding dense connection structures[17] inspired by[18].

All of these efforts are built on thoroughly annotated data sets such as COCO[19], PASCAL VOC2012[20], while the expenditure of annotating demands huge manpower and material resources. As for large-scale biomedical image segmentation tasks, idea of segmenting the whole image based on patch classification occupies the mainstream [4] while the Fully Convolutional Networks (FCN)-based model is frequently used in small-scale biomedical segmentation problems which requires for precisely labeled. In Ref.[21], the authors presented a unified framework for the prediction of tumor proliferation scores in breast histopathology and have won the Tumor Proliferation Assessment Challenge in MICCAI 2016. Ref. [22] propose a two-pathway network architecture that the full-connected pathway is used to concentrate on a small region around the pixel to classify, while the convolutional pathway looks at a wider region. A series of methods such as Atrous Convolution and Conditional Random Field (CRF) [23][24][25] have been widely used in natural semantic image segmentation problems and achieved surprising results. However, these methods are not appropriate for biomedical segmentation because the pooling layers leads to the reduction of output size (subsampled by 8 compared to the input), which makes large error in biomedical segmentation.

With the proposing of active learning framework in biomedical image classification problem[26][27], semi-supervised learning arouses researchers' extensive attention [8][28][29]. Wang and Shang [30] first combine active learning with deep learning, then Zhou [28] fine tune the CNNs in an incremental manner instead of repeatedly re-trained.

These methods do decrease the cost of annotation, while they still require a fraction of strongly annotated labels and the real-time supplemental labeling.

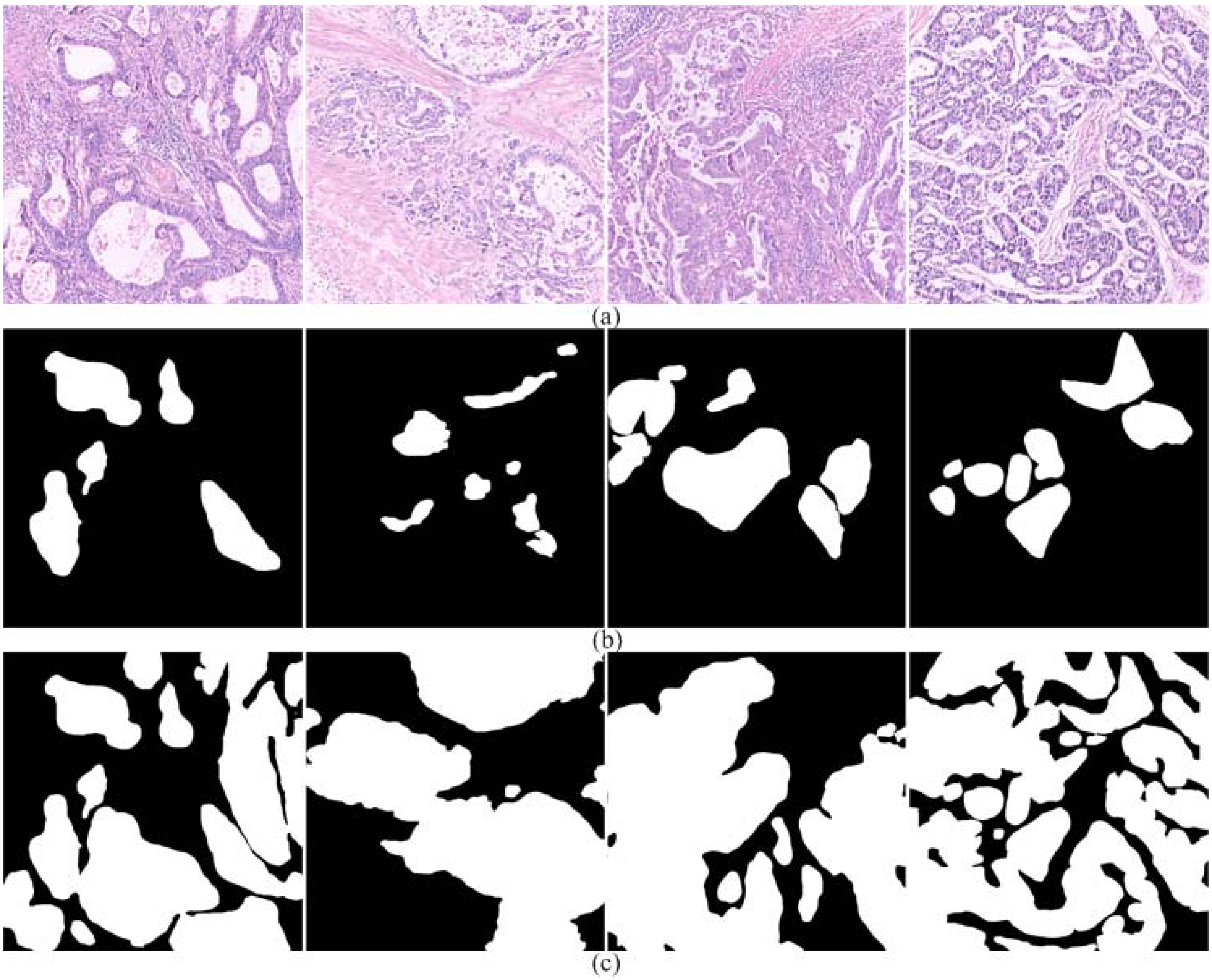

Fig. 1. Partial annotated images in the dataset, with 20% to 70% true positive area marked. (a) Biomedical images in training dataset (b) labels given by authorities (c) ground truth.

In this paper, we present a reiterative learning framework to dealing with partial and weakly labeled gastric cancer images. Each image in the dataset is partial marked as show in Fig. 1. Starting with training our FCN-based model on medium-sized patches selected by area threshold, problems of false negative samples are judiciously resolved. With the preprocessing of images, background and pipes formed by cancer are eliminated as to prevent false positive samples. Next, the whole training set is annotated by the patch model through overlapped region recast algorithm. We preserved those additional areas with high confidence as new labels, and presented a concept of reiterative learning framework and eventually reached mean Intersection Over Union (IOU) of 0.883 and mean accuracy of 91.09%. The remainder of our paper is organized as follows: we describe our approach in Sec.2. In Sec.3, details of experiments are given rigorously. The final conclusion is presented in Sec.4.

## 2 Approach

In this paper, we build a FCN-based convolutional network trained on medium-sized patches and eliminate the deviation between patches through Overlapped region forecast algorithm. Reiterative learning framework is utilized in order to promote the efficiency of the network. Although the above approach seems straightforward, we need to overcome a lot of challenges as to integrate patch model into our deep reiterative learning framework, our main contribution is as follows:

- We present an advisable training framework to train our network on partial annotated biomedical dataset and achieve the state of the art performance without any pre-trained and further manual annotation.
- We combine the patch-based approach with FCN-based model and present a new method to solve the biomedical segmentation problems.
- An overlapped region forecast algorithm was proposed to merge the predictions as to promote the performance of final results.

## 2.1 Regions of Interests Extraction and Preprocessing

The original image is 2048 * 2048 pixels with partial labeled, which incur to lots of false negative samples. We remove this obstacle through cutting the histopathology images into 512 * 512 pixels patches. First of all, the histopathology images were sampled at the sizes of 512 pixels and strides of 256 pixels. Then, patches with area threshold lower than 0.5 will be discarded, the area threshold $\mu$ is defined as follows:

$$\mu = \frac{C_A}{O_A} \quad (1)$$

where $C_A$ is the annotated area of cancer, $O_A$ is the area of overall patch. After these mentioned steps, a series of appropriate patches were acquired for the following steps.

As to eliminate the background and pipes generated by cancer cells, we provide an available algorithm using digital image processing technologies. Masks that represent the background and pipe is produced through binarization. The preprocessing image can be obtained through multiplying the mask's inversion by the patch label. As shown in Fig. 2, white background is effectively removed from patches while other regions are precisely remained.

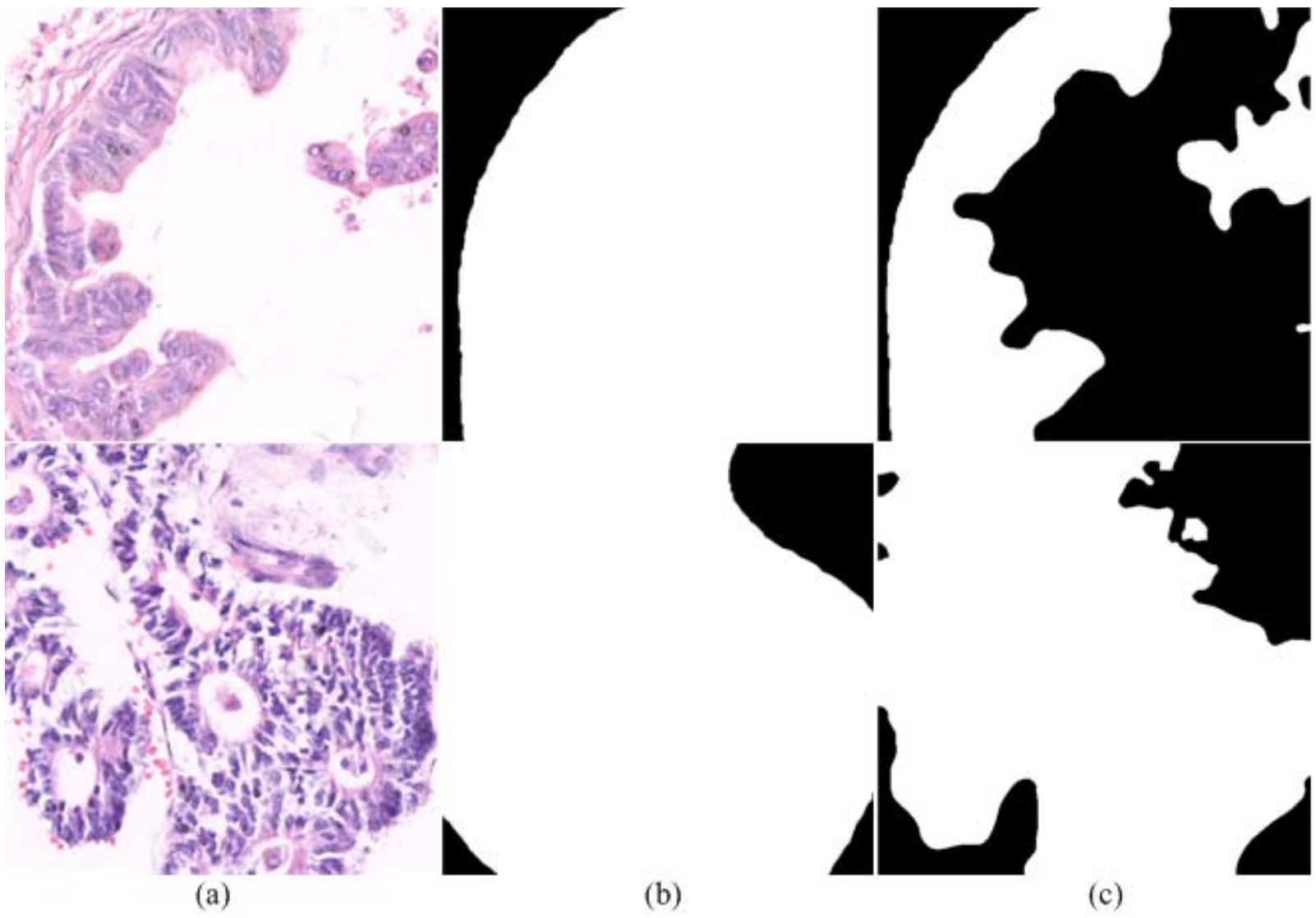

Fig. 2. (a) is the original images and (b) is the labels given by authorities. Background is judiciously removed in (c) while the cell matrix surrounded by CW (Cell Wall) is remained through Flood Fill Algorithm (FFA).

## 2.2 Patch-based Network and training strategies

Before the advent of FCN, patch-based method is widely used in biomedical image segmentation[31][32], these patch-based methods are efficient when facing huge size of whole slide biomedical image segmentation problems. However, when facing the medium-sized image segmentation task, it is invalid to use these patch-based methods as the limit to the image size.

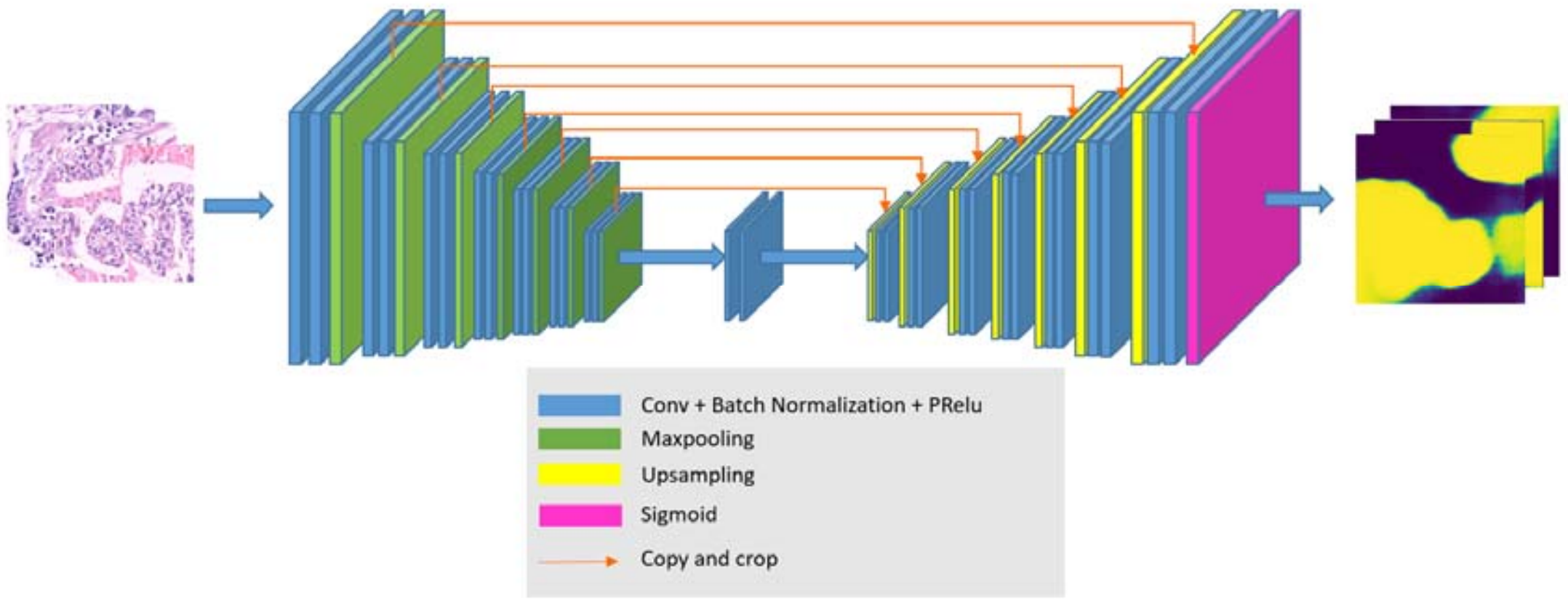

Fig. 3. Architecture of our model

In order to work out this problem, our system is composed of patch-based method and FCN-based model. The architectures of the proposed network are illustrated in Fig. 3, Parametric Relu is utilized in order to allow a small, non-zero gradient when the unit is not active by making the coefficient of leakage into a parameter that is learned along with the other neural network parameters [33]. As shown in Fig. 3, we trained patches through our network and obtained the probability heat maps. A proper loss function was defined which can be expressed as:

$$\text{loss} = loss_{bce} + loss_{dice} \quad (2)$$

where $loss_{bce}$ is the binary cross entropy loss, supposed we have N samples with each sample labeled by $n = 1,2, \ldots \ldots N$. The binary cross entropy loss function is then given by:

$$\text{L(w)} = \frac{1}{N}\sum_{n=1}^{N} H(p_n, q_n) = -\frac{1}{N}\sum_{n=1}^{N}[y_n log\widehat{y_n} + (1 - y_n)\log(1 - \widehat{y_n})] \quad (3)$$

The dice coefficient loss $loss_{dice}$ can be computed as:

$$loss_{dice} = 1 - 2 \times \frac{|GT \cap SR| + smooth}{|GT| + |SR| + smooth} \quad (4)$$

where GT is the real ground truth (all of the discernible parts of glands or crypts have been labeled), SR represents segmentation results. As the experiments show, our new model can achieve brilliant performance with training on patches.

## 2.3 Overlapped Region Forecast (ORF) and Post-processing

An advisable method of merging patches is presented in this section, which eliminate the obstacle of predicting original images. Due to the characteristics of the cancer gland morphology, the edge part of the patches is easier to present coarse predictions, for single cancer gland might be sampled to different patches while those white pipes within the gland is similar to the backgrounds.

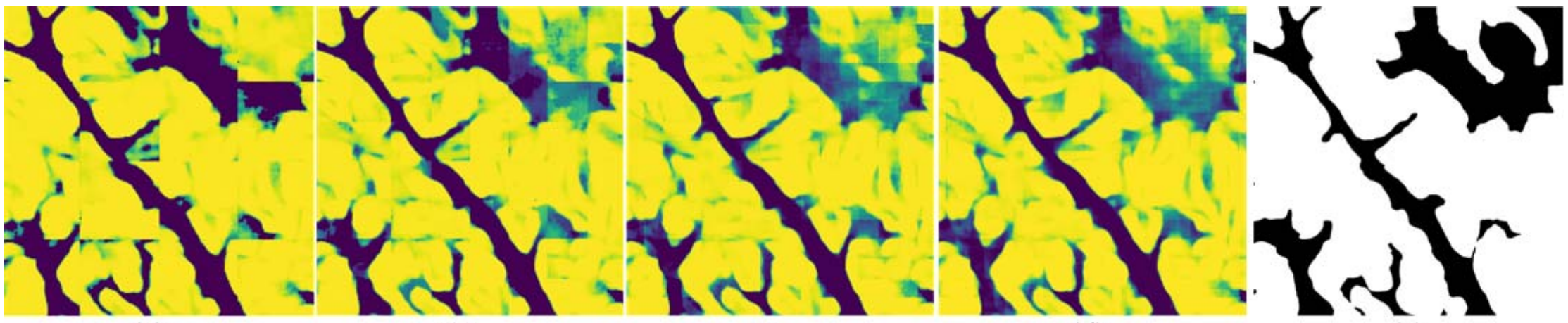

(a) (b) (c) (d) (e)

Fig. 4. Overlapped region forecast and post—precession obtained by different strides. (a)-(d) are the belief maps corresponding to the strides of 512,256,128 and 64. (e) is the ground truth annotated by experts.

To fix this hurdle, we present a method named overlapped region forecast, which calculates the probability of cancer at pixel-level and preserves the results into a list called gallery. After the gallery of an image is obtained, the probability value at each pixel in the patches is added together, depending on its initial position in the original image. The value of pixel $x_{ij}$ can be expressed as:

$$x_{ij} = \frac{m_{ij}}{a_i * a_j} \tag{5}$$

where $m_{ij}$ is the value of the i th row and j th column in the matrix after the superposition and $a_i, a_j$ is calculated through:

$$a_n = \begin{cases} ceil\left(\frac{n}{stride}\right) & else \\ \frac{s_p}{stride} & ceil\left(\frac{n}{stride}\right) \geq \frac{s_p}{stride} \end{cases} \tag{6}$$

The number of patches $N_g$ in each gallery is described as:

$$N_g = (\frac{s_{ori} - s_p}{stride} + 1)^2 \tag{7}$$

where $s_{ori}$ is the size of original histopathology images, $s_p$ is the size of patches, stride is the submultiple of patch size. When the stride is set to 512, the final prediction is straightly merged by 16 patches without overlapped region. We also use an 11*11 average filter to smooth the edges of the cancer area. The prediction results with different strides are illustrated in Fig. 4, which indicates that the ORF do make the model pay more attention to the distinction between the edge and the details.

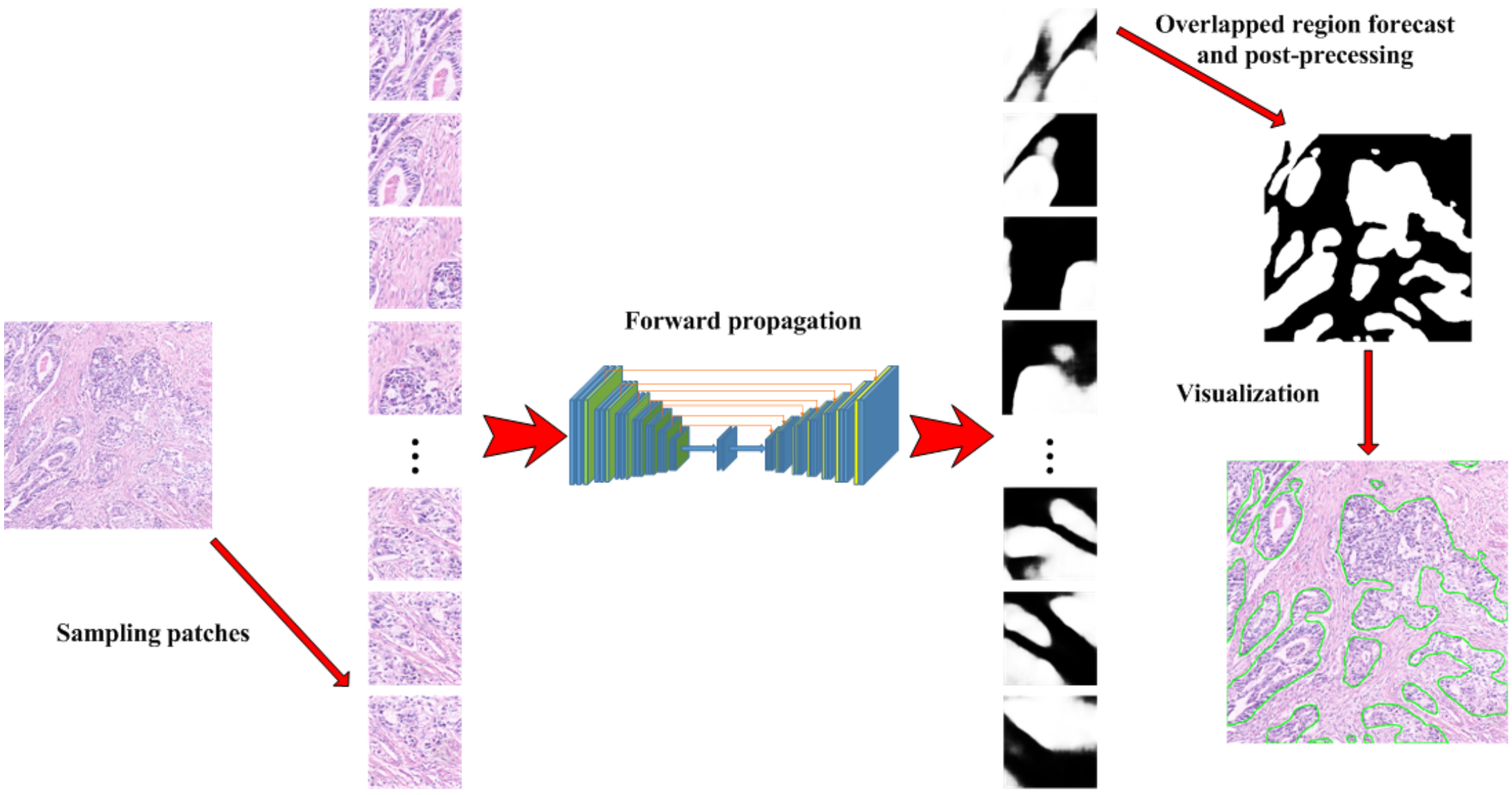

Fig. 5. Overall validation steps of our model.

Due to the rules of segmentation in this challenge, necrosis or pipes in glands are desired to be annotated as negative samples. Flood fill (seed fill), is an algorithm that determines the area connected to a given node in a multi-dimensional array. We followed this procedure to eliminate those “false negative” areas within the glands. The entire evaluation process is demonstrated in Fig. 5.

## 2.4 Reiterative Learning Framework

With the patch-based training method proposed, problems caused by weakly annotated have been dexterously avoided. However, these attempts are not sufficient to achieve excellent results, errors caused by weakly labeled (shown in Fig. 6) still leads to the misjudgment of the model.

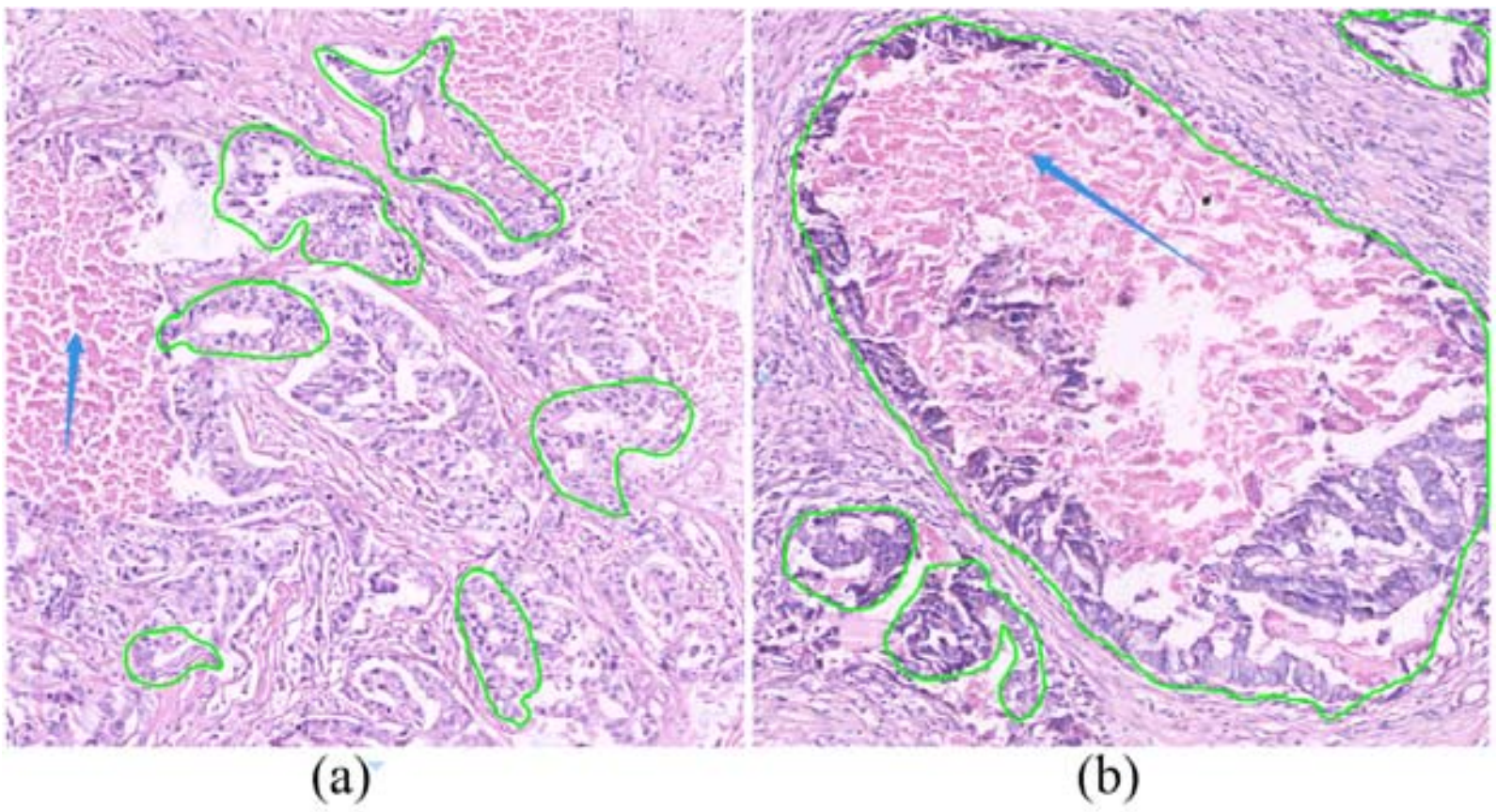

Fig. 6. Errors caused by weak annotations. The necrotic regions (pointed by blue arrows) in (a) and (b) have the opposite annotations.

To solve this problem, a novel training strategy named reiterative learning is proposed, which is demonstrated in Fig. 7. We follow this procedure to train our model and achieve excellent performance in the challenge.

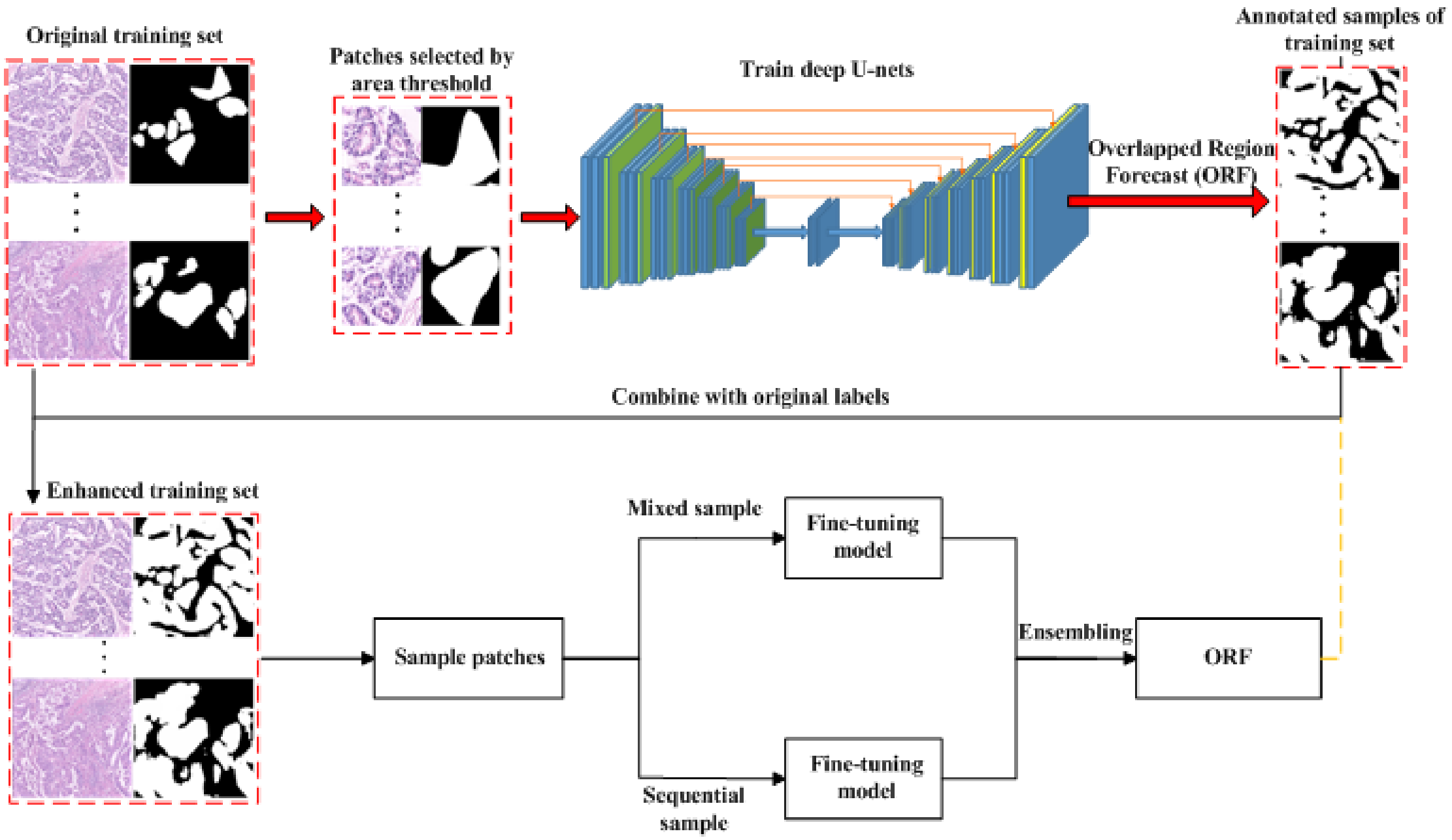

Fig. 7. Illustrating our overall reiterative learning pipeline.

When training on patch images, we acquire a powerful segmentation model which can achieve the outstanding performance on patches. Rather than directly predicting the test images, the model was used to perform overlapped region forecast (ORF) on the original training dataset and reserve the positive regions with a high confidence level of 0.7. To supplement those areas labeled by the authorities, we combining the original labels with high-confidence-level annotations. This allows the original training data set to be expanded, and most of the positive area has been credible labeled. As we know, dicing by area results in a reduction of false-negative samples, whereas sequential dicing increases the number of true-negative samples, so we slice the patches in these two different ways and acquire two new data sets named mix-patches set and sequential-patches set. The precision and recall can be quantified as:

$$\text{precision} = \frac{TP}{TP+FP} \tag{8}$$

$$\text{recall} = \frac{TP}{TP+FN} \tag{9}$$

Due to the characteristics of the two patch datasets described above, the model trained on the mixed-patches has a lower FN, corresponding to a higher recall rate, whereas the model trained on the sequential-patches has a lower FP resulting in a higher accuracy, as shown in Fig. 8.

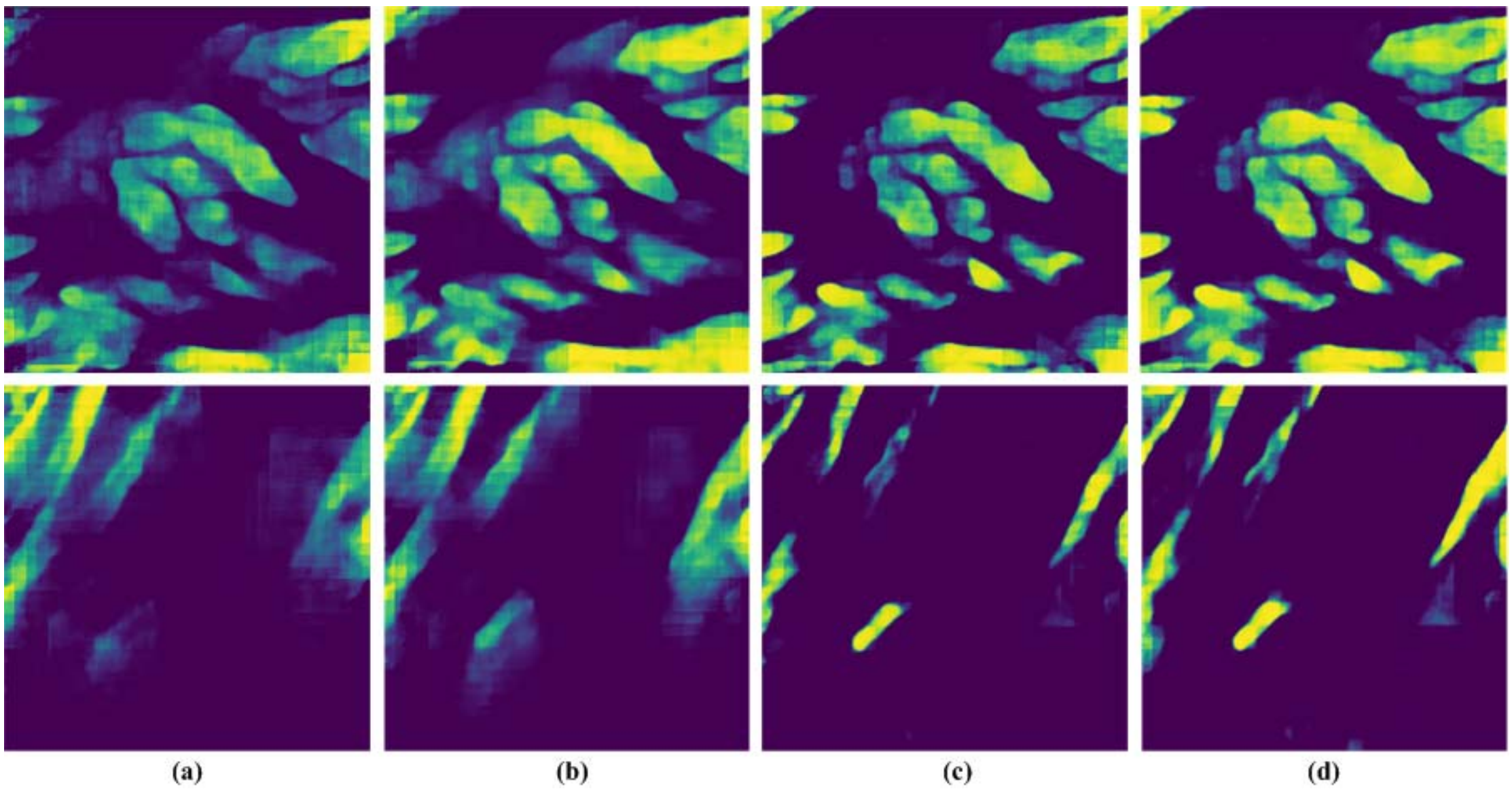

Fig. 8. Different belief maps of each stage during the reiterative learning. (a) Our patch model (b)"1st review" (training on mix-patch dataset) (c) "1st review" (training on sequential-patch dataset) (d) "2nd review"

Next, we separately fine-tune the models on two datasets and ensemble them altogether. Annotated through the ensemble model, the enhanced original label set is acquired. We refer to this process as the "first review," and the quantitative results will be demonstrated in Sec.4. According to this strategy, we carry out the "second review" and achieve excellent performance in the challenge.

## 3 Experiments

We evaluate our framework on gastric tumor segmentation data set, which was provided by 2017 China Big Data & Artificial Intelligence Innovation and Entrepreneurship Competition. The whole training data is a set of 1400 biomedical images (2048 * 2048 pixels) with weakly annotated and the test set consists of 500 precisely labeled images.

Table 1. Metrics of different strides in ORF, generating by our patch model

| Stride | Mean IOU | Mean accuracy | Time consuming per image/s |
|---|---|---|---|
| 512 | 82.73 | 86.58 | 5 |
| 256 | 83.91 | 87.93 | 7 |
| 128 | 85.19 | 88.67 | 11 |
| 64 | 85.28 | 88.75 | 25 |

First, we assess the impact of overlapped region forecast algorithm. The quantitative results in Table 1 show that the effect of different strides, where stride of 512 indicates that we just merge the predictions straightly without ORF. As it can be seen in Table1, applying ORF with 64 strides can achieve the best results while it requires a great deal of time consuming. The stride we used during the experiment was 128 due to its satisfying capability.

Herein, we evaluate our models on the test data set and present the results obtained by each stage in the reiterative learning procedures. In Table2, four metrics from common biomedical segmentation problems are reported: precision, recall, accuracy and the mean IOU (we invert the label and calculate the IOU if the image is pure negative). Table 2 gives the results of our model in different stages and the performance of other teams, which indicates the efficiency of reiterative learning framework.

Table 2. Quantitative results of our models and other teams ("*" stands for unknown, as the authority only revealed the mean IOU.)

| Model | Mean precision | Mean recall | Mean accuracy | Mean IOU |
|---|---|---|---|---|
| Patch model | 86.64 | 86.65 | 88.75 | 85.28 |
| 1st review (training on sequential-patches) | 90.68 | 84.95 | 89.64 | 86.46 |
| 1st review (training on mix-patches) | 87.45 | 87.37 | 89.35 | 86.21 |
| 1st ensemble model | 89.80 | 85.77 | 89.71 | 86.67 |
| 2nd review (training on sequential-patches) | 91.23 | 87.74 | 90.46 | 87.53 |
| 2nd review (training on mix-patches) | 90.71 | 88.39 | 90.27 | 87.46 |
| Up (our 2nd ensemble model) | 91.11 | 88.05 | 91.09 | 88.31 |
| B&D (2nd team in the ranking list) | * | * | * | 87.51 |
| PA Tech (3rd team in the ranking list) | * | * | * | 87.06 |

As shown in Table 2, the performance of the model has been apparently improved after the first iterative learning. Model trained on mix-patch dataset has high recall, while model trained on sequential-patch dataset has high precision. After ensemble those two models and do the "second review", we reached the state of the art performance in the 2017 China Big Data & Artificial Intelligence Innovation and Entrepreneurship Competitions[2]. Turning to qualitative results, we provide visual comparisons between our model and the ground truth shown in Fig. 9.

[2] The ranking list is provided in http://www.datadreams.org/race-list-3.html

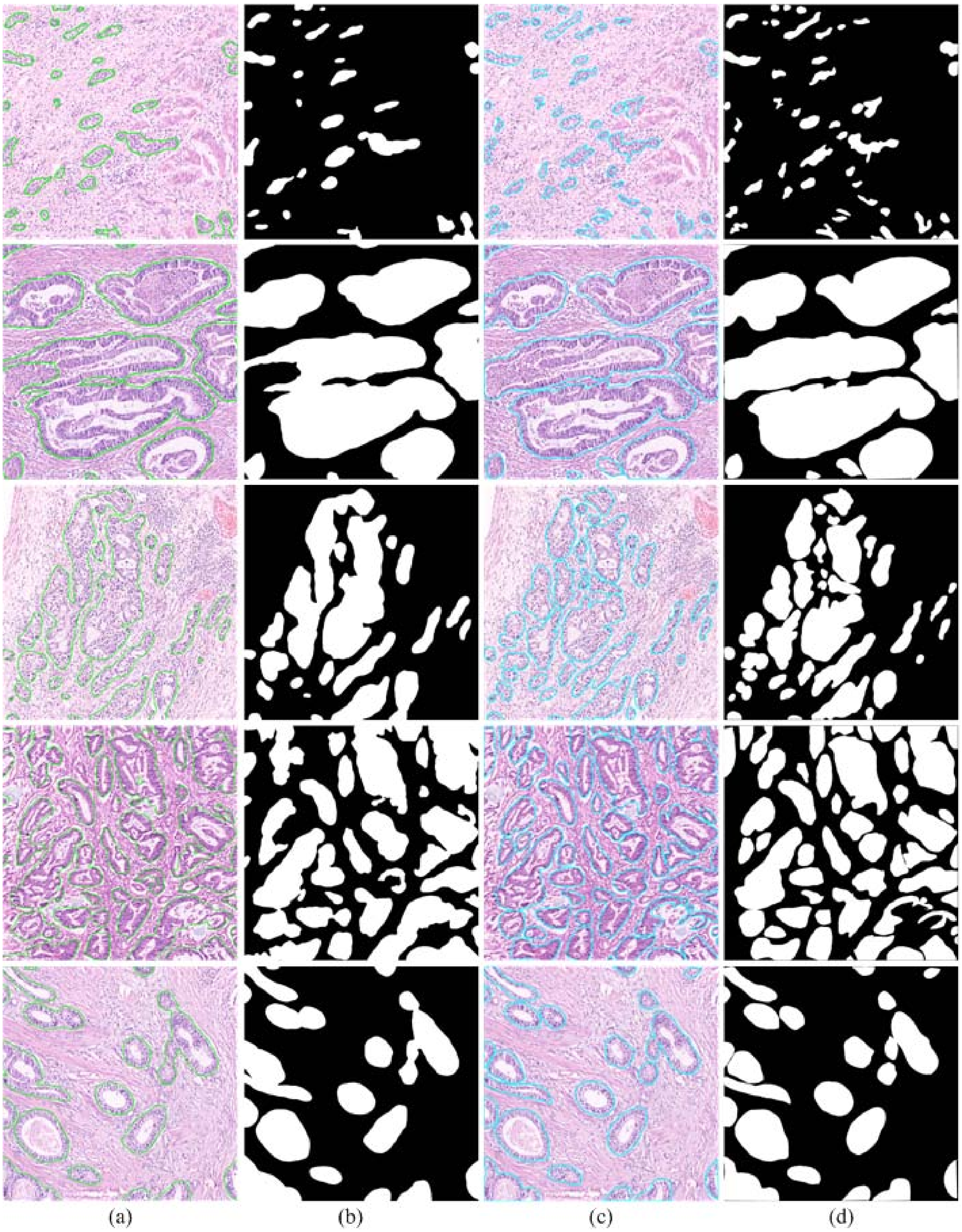

Fig. 9. Qualitative segmentation results on the data set. (a) Visualizations of our model (b) Labels of our model (c) Visualizations of the ground truth (d) Labels of the ground truth

# 4 Conclusion

In this paper, we studied the use of partial annotation in training a state of art biomedical image segmentation model. The proposed framework "reiterative learning" allows the network to achieve outstanding performance with training on the weakly annotated dataset without any manual annotation. Specifically, we combine the patch-based method with FCN-based method and present an overlapped region forecast algorithm to optimize the final results, avoiding the risk of over-fitting in the first few steps of reiterative learning. Through these efficient approaches, a mean intersection over union Coefficient (IOU) of 0.883 and mean accuracy of 91.09% on the partial labeled data set was acquired. These strategies have made our model achieved the supervised learning standard and won the championship of the 2017 China Big Data & Artificial Intelligence Innovation and Entrepreneurship Competitions. The proposed approach could greatly reduce the cost of labeling and the limitations of histopathology research, which will further promote the research of biomedical histopathology segmentation problems.

However, as shown in Fig. 9, though our model has generally yielded satisfactory results, improvements are still required in the following details:

(1) Ability of our model to recognize small cancerous tissues needs to be further improved.

(2) Over-fitting may occur when iterative learning too many times.

## Acknowledgements

This work was supported in part by the National Nature Science Foundation of China (NSFC 61673163), Hunan provincial natural science foundation of China (2016JJ3045).

## References

[1]Ronneberger, Olaf, P. Fischer, and T. Brox. "U-Net: Convolutional Networks for Biomedical Image Segmentation." International Conference on Medical Image Computing and Computer-Assisted Intervention Springer, Cham, 2015:234-241.
[2] Chen, Hao, et al. "DCAN: Deep Contour-Aware Networks for Accurate Gland Segmentation." *Computer Vision and Pattern Recognition* IEEE, 2016:2487-2496.
[3] Chen, Hao, et al. "Deep contextual networks for neuronal structure segmentation." *Thirtieth AAAI Conference on Artificial Intelligence* AAAI Press, 2016:1167-1173.
[4] Esteva, A, et al. "Corrigendum: Dermatologist-level classification of skin cancer with deep neural networks." *Nature* 542.7639(2017):115.
[5] Papandreou, George, et al. "Weakly-and Semi-Supervised Learning of a Deep Convolutional Network for Semantic Image Segmentation." *IEEE International Conference on Computer Vision* IEEE Computer Society, 2015:1742-1750.
[6] Hong, Seunghoon, H. Noh, and B. Han. "Decoupled Deep Neural Network for Semi-supervised Semantic Segmentation." (2015):1495-1503.
[7] Jain, Suyog Dutt, and K. Grauman. "Active Image Segmentation Propagation." *Computer Vision and Pattern Recognition* IEEE, 2016:2864-2873.
[8] Yang, Lin, et al. *Suggestive Annotation: A Deep Active Learning Framework for Biomedical Image Segmentation. Medical Image Computing and Computer-Assisted Intervention − MICCAI 2017*. 2017.
[9] Drozdzal, Michal, et al. "The Importance of Skip Connections in Biomedical Image Segmentation." (2016):179-187.
[10] Wang, Guotai, et al. "Automatic Brain Tumor Segmentation using Cascaded Anisotropic Convolutional Neural Networks." (2017).
[11] Yan, Zhicheng, et al. "Combining the Best of Convolutional Layers and Recurrent Layers: A Hybrid Network for Semantic Segmentation." (2016).

[12] Visin, Francesco, et al. "ReSeg: A Recurrent Neural Network-Based Model for Semantic Segmentation." *Computer Vision and Pattern Recognition Workshops* IEEE, 2016:426-433.
[13] Visin, Francesco, et al. "ReSeg: A Recurrent Neural Network-Based Model for Semantic Segmentation." *Computer Vision and Pattern Recognition Workshops* IEEE, 2016:426-433.
[14] Milletari, Fausto, N. Navab, and S. A. Ahmadi. "V-Net: Fully Convolutional Neural Networks for Volumetric Medical Image Segmentation." *Fourth International Conference on 3d Vision* IEEE, 2016:565-571.
[15] Özgün Çiçek, et al. "3D U-Net: Learning Dense Volumetric Segmentation from Sparse Annotation." *International Conference on Medical Image Computing and Computer-Assisted Intervention* Springer, Cham, 2016:424-432.
[16] Abdulnabi, Abrar H, S. Winkler, and G. Wang. "Beyond Forward Shortcuts: Fully Convolutional Master-Slave Networks (MSNets) with Backward Skip Connections for Semantic Segmentation." (2017).
[17] Jégou, Simon, et al. "The One Hundred Layers Tiramisu: Fully Convolutional DenseNets for Semantic Segmentation." (2016):1175-1183.
[18] Huang, Gao, Z. Liu, and K. Q. Weinberger. "Densely Connected Convolutional Networks." *CVPR* 2016.
[19] Lin, Tsung Yi, et al. "Microsoft COCO: Common Objects in Context." 8693(2014):740-755.
[20] Everingham, Mark, et al. "The Pascal, Visual Object Classes (VOC) Challenge." *International Journal of Computer Vision* 88.2(2010):303-338.
[21] Paeng, Kyunghyun, et al. "A Unified Framework for Tumor Proliferation Score Prediction in Breast Histopathology." (2016).
[22] Havaei, Mohammad, et al. "Brain tumor segmentation with Deep Neural Networks." *Medical Image Analysis* 35(2017):18-31.
[23] LiangChieh Chen, et al. "Semantic Image Segmentation with Deep Convolutional Nets and Fully Connected CRFs." *Computer Science* 4 (2014):357-361.
[24] Chen, Liang Chieh, et al. "DeepLab: Semantic Image Segmentation with Deep Convolutional Nets, Atrous Convolution, and Fully Connected CRFs." *IEEE Transactions on Pattern Analysis & Machine Intelligence* PP. 99 (2016):1-1.
[25] Chen, Liang Chieh, et al. "Rethinking Atrous Convolution for Semantic Image Segmentation." (2017).
[26] Rajan, Suju, J. Ghosh, and M. M. Crawford. "An Active Learning Approach to Hyperspectral Data Classification." *IEEE Transactions on Geoscience & Remote Sensing* 46.4(2015):1231-1242.
[27] Zhou, Zongwei, et al. "Fine-tuning Convolutional Neural Networks for Biomedical Image Analysis: Actively and Incrementally *." *The IEEE Conference on Computer Vision and Pattern Recognition* IEEE, 2017.
[28] Huang, G., et al. "Semi-supervised and unsupervised extreme learning machines. " *IEEE Transactions on Cybernetics* 44.12(2017):2405-2417.
[29] Papandreou, George, et al. "Weakly- and Semi-Supervised Learning of a DCNN for Semantic Image Segmentation." (2015):1742-1750.
[30] D. Wang and Y. Shang. A new active labeling method for deep learning. In 2014 International Joint Conference on Neural Networks (IJCNN), pages 112–119, July 2014
[31] Coupé, P, et al. "Patch-based segmentation using expert priors: application to hippocampus and ventricle segmentation." *Neuroimage* 54. 2(2011):940-954.
[32] Wang, Dayong, et al. "Deep Learning for Identifying Metastatic Breast Cancer." (2016).
[33] He, Kaiming, et al. "Delving Deep into Rectifiers: Surpassing Human-Level Performance on ImageNet Classification." *IEEE International Conference on Computer Vision* IEEE Computer Society, 2015:1026-1034.